\documentclass[11pt,a4paper]{article}
\usepackage[hyperref]{acl2019}
\usepackage{times}
\usepackage{latexsym}
\usepackage{url}
\usepackage{times}
\usepackage{latexsym}
\usepackage{amsmath}
\usepackage{amsfonts}
\usepackage{microtype}
\usepackage{xfrac}
\usepackage{url}
\usepackage{mathtools}
\usepackage{tikz}
\usepackage{caption}
\usetikzlibrary{shapes.geometric}
\usetikzlibrary{automata, positioning, arrows}
\usepackage{kotex}
\usepackage{booktabs}
\usepackage[T1]{fontenc}
\usepackage[safe]{tipa}
\usepackage{adjustbox}
\usepackage{xspace}
\usepackage[normalem]{ulem}
\usepackage{subfig}
\usepackage{pifont}

\usepackage{xcolor}
\usepackage{algorithm}
\usepackage[noend]{algpseudocode}

\usepackage{cleveref}
\crefname{section}{\S}{\S\S}
\Crefname{section}{\S}{\S\S}
\crefname{table}{Tab.}{}
\crefname{figure}{Fig.}{}
\crefname{algorithm}{Alg.}{}
\crefname{equation}{eq.}{}
\crefname{appendix}{App.}{}
\crefformat{section}{\S#2#1#3}  %

\usepackage{todonotes}

\newcommand{\softmax}{\text{softmax}\xspace}

\newcommand{\relu}{\text{ReLU}\xspace}
\newcommand{\lstm}{\text{LSTM}\xspace}
\newcommand{\yy}{\boldsymbol{y}}
\newcommand{\xx}{\boldsymbol{x}}
\newcommand{\ttt}{\mathbf{t}}
\newcommand{\cc}{\mathbf{c}}
\newcommand{\al}{\boldsymbol{a}}

\newcommand{\NN}{\mathbf{f}} %
\newcommand{\dech}{\mathbf{h}^{\textit{d}}}
\newcommand{\ench}{\mathbf{h}^{\textit{e}}}
\newcommand{\tagh}{\mathbf{h}^{\textit{t}}}
\newcommand{\dece}{\mathbf{e}^{\textit{d}}}

\newcommand{\tage}{\mathbf{e}^{\textit{t}}}

\newcommand{\eos}{\textsc{eos}}

\newcommand{\calA}{\mathcal{A}}

\newcommand{\Sigmay}{\Sigma_{\texttt{y}}}
\newcommand{\Sigmax}{\Sigma_{\texttt{x}}}
\newcommand{\Sigmat}{\Sigma_{\texttt{t}}}

\newcommand{\mathbbm}[1]{\text{\usefont{U}{bbm}{m}{n}#1}} %

\usepackage{inconsolata}

\aclfinalcopy %
\def\aclpaperid{548} %

\title{Exact Hard Monotonic Attention for Character-Level Transduction}

\author{Shijie Wu$^{\textrm{\normalfont \textschwa}}$ \and Ryan Cotterell$^{\textrm{\normalfont \textschwa,\textipa{H}}}$ \\
${}^{\textrm{\textschwa}}$Department of Computer Science, Johns Hopkins University \\
${}^{\textrm{\textipa{H}}}$Department of Computer Science and Technology, University of Cambridge  \\
{\tt shijie.wu@jhu.edu} \quad {\tt rdc42@cam.ac.uk}
}

\date{}

\begin{document}
\maketitle
\begin{abstract}
Many common character-level, string-to-string transduction tasks, e.g., grapheme-to-phoneme conversion and morphological
inflection, consist almost exclusively of monotonic transductions. 
However, neural sequence-to-sequence models that use non-monotonic soft attention often outperform popular monotonic models. 
In this work, we ask the following question: Is monotonicity really a helpful inductive bias for these tasks? 
We develop a hard attention
sequence-to-sequence model that enforces strict monotonicity and learns a latent alignment jointly while learning to transduce. With the help of dynamic programming, we are able to compute the exact
marginalization over all monotonic alignments. 
Our models achieve state-of-the-art performance on morphological inflection. Furthermore, we find
strong performance on two other character-level
transduction tasks. 
Code is available at \url{https://github.com/shijie-wu/neural-transducer}.\looseness=-1
\end{abstract}

\section{Introduction}
Many tasks in natural language processing can be treated as character-level,
string-to-string transduction. The current dominant method is the neural sequence-to-sequence model with soft attention
\citep{bahdanau+al-2014-nmt,Luong2015EffectiveAT}. This method has achieved state-of-the-art results in a plethora of tasks, for example,
grapheme-to-phoneme conversion \cite{DBLP:conf/interspeech/YaoZ15}, named-entity
transliteration \cite{rosca2016sequence} and morphological inflection generation
\cite{W16-2002}. While soft attention is very similar to a traditional alignment between the source characters and target characters in some regards, it does not explicitly model a distribution over alignments. On the other hand, neural
sequence-to-sequence models with hard attention
are analogous to the classic IBM models for machine translation, which \emph{do} model the alignment distribution explicitly \cite{brown1993mathematics}.\looseness=-1

\begin{figure*}
\centering
\includegraphics[width=2\columnwidth]{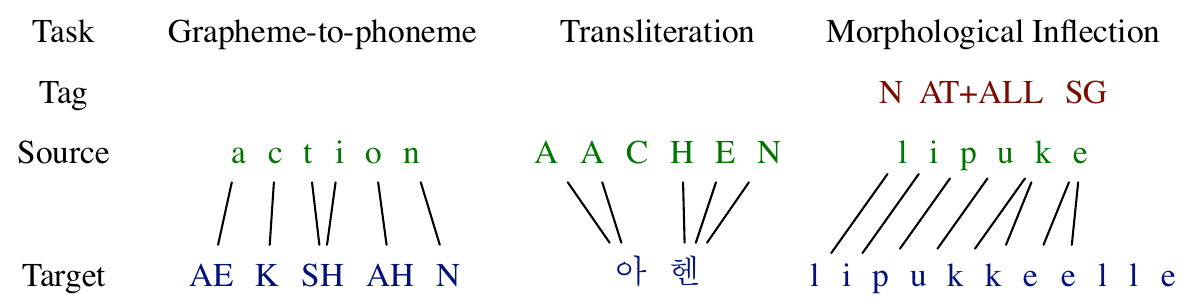}
\caption{Example of source and target string for each task. Tag guides transduction in morphological inflection.}
\label{fig:example}
\vspace{-.6cm}
\end{figure*}

The standard versions of soft and hard attention are \emph{non}-monotonic. 
However, if we look at the data in grapheme-to-phoneme
conversion, named-entity transliteration, and morphological inflection (examples are shown in \cref{fig:example}), we see that the tasks require almost exclusively \emph{monotonic} transduction. 
Yet, counterintuitively, the state of the art in high-resource morphological inflection is held by non-monotonic models \citep{cotterell-conll-sigmorphon2017}! Indeed, in a recent controlled experiment, \newcite{wu2018hard} found non-monotonic models (with
either soft or hard attention) outperform popular monotonic models \cite{aharoni-goldberg:2017:Long} in the three above-mentioned
tasks. However, the inductive bias of monotonicity, if correct, should help learn a better model or, \emph{at least},
learn the same model.\looseness=-1 

In this paper, we hypothesize that the underperformance
of monotonic models stems from the lack of joint training
of the alignments with the transduction.
Generalizing the 
model of \newcite{wu2018hard} to enforce monotonic alignments,
we show that, for all three tasks considered, monotonicity is a good inductive bias and
jointly learning a monotonic alignment improves performance. We provide an exact, cubic-time dynamic-programming inference
algorithm to compute the log-likelihood and an approximate
greedy decoding scheme.
Empirically, our results indicate that, rather than
the pipeline systems of \newcite{aharoni-goldberg:2017:Long} and \newcite{makarov2017align},
we should jointly train monotonic alignments with the transduction model, and, indeed, we
set the single-model state of the art on the task of morphological inflection.\footnote{The current state of the art for morphological inflection is held by ensemble systems like parsing and other structured prediction tasks. We present the new \emph{best individual} system.\looseness=-1} \looseness=-1

\section{Hard Attention}\label{sec:hard-attention}

\subsection{Preliminary}
We assume the source string $\xx \in \Sigmax^*$ and the target string $\yy \in
\Sigmay^*$ are drawn from finite vocabularies $\Sigmax = \{x_1, \ldots, x_{|\Sigmax|}\}$
and $\Sigmay = \{ y_1, \ldots, y_{|\Sigmay|} \}$, respectively. 
In tasks where the tag is provided, i.e., labeled transduction \cite{zhou-neubig:2017:Long}, we denote the tag as an ordered set $\ttt \in \Sigmat^*$ drawn from a finite tag vocabulary
$\Sigmat = \{ t_1, \ldots, t_{|\Sigmat|} \}$. We define the set $\calA =\{1,
\ldots, |\xx|\}^{|\yy|}$ to be set of all non-monotonic alignments from $\xx$ to $\yy$ where
an alignment aligns each target character $y_i$ to exactly one source character
in $\xx$.\footnote{We write $\mathcal{A}$ in
the remainder with $\xx$ and $\yy$ implicit.}
In other words, it allows zero-to-one\footnote{Zero in the sense of a non-character like \texttt{BOS} or \texttt{EOS}} or many-to-one alignments
between $\xx$ and $\yy$. For an $\al \in \calA$, $A_i = a_i$ refers to the event that
$y_i$ is aligned to $x_{a_i}$, which are the $i^\text{th}$ character of $\yy$ and the
${a_i}^\text{th}$ character of $\xx$, respectively.
In general, we will shorten the expression $A_i = a_i$ to $a_i$ for brevity.\looseness=-1

\subsection{0$^\text{th}$-order Hard Attention}

Hard attention was first introduced to
the literature by \newcite{xu2015show}.
We, however, follow \newcite{wu2018hard} and
use a tractable variant of hard attention
and model the probability of a target 
string $\yy$ given an input string $\xx$
as follows

\begin{align}
&p(\yy \mid \xx) = \sum_{\al \in \calA} p(\yy, \al \mid \xx) \label{eq:0th-order} \\
&= \underbrace{\sum_{\al \in \calA} \prod_{i=1}^{|\yy|} p(y_i \mid a_i, \yy_{< i}, \xx)\, p(a_i \mid \yy_{< i}, \xx)}_{\textit{exponential number of terms}} \nonumber  \\
&= \underbrace{\prod_{i=1}^{|\yy|} \sum_{a_i=1}^{|\xx|} p(y_i  \mid a_i, \yy_{< i}, \xx)\,p(a_i \mid \yy_{< i}, \xx )}_{\textit{polynomial number of terms}} \nonumber
\end{align}
where we show how one can rearrange the terms to compute the function in polynomial time.\looseness=-1

The model above is exactly an 0$^\text{th}$-order neuralized hidden Markov model (HMM). Specifically, $p(y_i  \mid a_i, \yy_{< i}, \xx)$ can be regarded as an
emission distribution and $p(a_i \mid \yy_{< i}, \xx )$ can be regarded as a
transition distribution, which \emph{does not} condition on the previous alignment. 
Hence, we will
refer to this model as 0$^\text{th}$-order hard attention. The likelihood can be computed in ${\cal O}(|\xx| \cdot |\yy| \cdot |\Sigma_{\texttt{y}}|)$ time.\looseness=-1

\subsection{1$^\text{st}$-order Hard Attention}
To enforce monotonicity, hard attention with conditionally independent alignment decisions is not
enough: The model needs to know the previous alignment position when determining the current alignment position. Thus, we allow the transition distribution to condition on the previous alignment $p(a_i \mid a_{i-1}, \yy_{< i}, \xx )$ and it becomes a 1$^\text{st}$-order neuralized HMM. We display this model as a graphical model in \cref{fig:graphical-model}. 
We will refer to it as 1$^\text{st}$-order hard attention.
Generalizing the $0^\text{th}$-order model, we define the $1^\text{st}$-order extension as follows
\begin{align}
&p(\yy \mid \xx) = \sum_{\al \in \calA} p(\yy, \al \mid \xx) \label{eq:1st-order} \\
&= \underbrace{\sum_{\al \in \calA} \prod_{i=1}^{|\yy|} p(y_i \mid a_i, \yy_{< i}, \xx)\, p(a_i \mid a_{i-1}, \yy_{< i}, \xx)}_{\textit{exponential number of terms}} \nonumber \\
&= \underbrace{\prod_{i=1}^{|\yy|} \sum_{a_{i-1}=1}^{|\xx|} \sum_{a_i=1}^{|\xx|} p(y_i \mid a_i)\,p(a_i \mid a_{i-1})\,\alpha(a_{i-1})}_{\textit{polynomial number of terms}}   \nonumber
\end{align}
where $\alpha(a_{i-1})$ is the forward probability,
calculated using the forward algorithm \cite{rabiner1989tutorial} with $\alpha(a_0,y_0) = 1$, and $p(a_1 \mid a_0) =
p(a_1\mid \texttt{<BOS>}, \xx)$ is the initial alignment distribution. For
simplicity, we drop $\yy_{< i}$ and $\xx$
in $p(y_i \mid a_i)$ and $p(a_i \mid a_{i-1})$.
For completeness, we include the
recursive definition of the forward probability:
\begin{align}
\alpha(a_{i}) &= p(y_i \mid a_i)\sum_{a_{i-1}=1}^{|\xx|} p(a_i \mid a_{i-1})\,\alpha(a_{i-1}) \nonumber \\
\alpha(a_1) &= p(y_1 \mid a_1)\,p(a_1\mid a_0)\,\alpha(a_0)
\end{align}
Thus, computation of the likelihood in our $1^\text{st}$-order hard attention model is
${\cal O}(|\xx|^2 \cdot |\yy| \cdot |\Sigma_\texttt{y}|)$ by the dynamic program given in the paper.

Decoding at test time, however, is hard and we resort to a greedy scheme,
described in \cref{alg:decode}. 
To see why it is hard, note that the dependence on $\yy_{<i}$ means that we have a neural language model scoring the target string
as it is being transduced. 
The dependence is unbounded so there is no dynamic program that allows for efficient computation.\looseness=-1

\begin{algorithm*}
\caption{Greedy decoding. ($N$ is the maximum length of the target string.)}\label{alg:decode}
\begin{algorithmic}[1]
\For{$i = 1,\ldots,N$}
\If {$i = 1$}
\State 
$y^*_i = \mathop{\text{argmax}}_{y_i} \sum^{|\xx|}_{a_i=1} p(y_i\mid a_i) p(a_i\mid a_{i-1})\,\alpha(a_0)$ \Comment{Greedy decoding}
\State $\alpha(a_1) = p(y^*_1 \mid a_1)\,p(a_1\mid a_0)\,\alpha(a_0)$ \Comment{Forward probability}
\Else
\State $y^*_i = \mathop{\text{argmax}}_{y_i} \sum^{|\xx|}_{a_i=1} p(y_i\mid a_i) \sum^{|\xx|}_{a_{i-1}=1} p(a_i\mid a_{i-1})\,\alpha(a_{i-1})$ \Comment{Greedy decoding}
\State $\alpha(a_{i}) = p(y^*_i \mid a_i)\sum_{a_{i-1}=1}^{|\xx|} p(a_i \mid a_{i-1})\,\alpha(a_{i-1})$ \Comment{Forward probability}
\EndIf
\If {$y^*_i = \texttt{EOS}$}
\State $\textbf{return}\, \yy^*$
\EndIf
\EndFor
\end{algorithmic}
\end{algorithm*}

\begin{figure}
\centering
\begin{adjustbox}{width=.9\columnwidth}
\begin{tikzpicture}
\tikzstyle{main}=[circle, inner sep = 0cm, minimum size = 10mm, thick, draw =black!80, node distance = 10mm]
\tikzstyle{blank}=[circle, inner sep = 0cm, minimum size = 10mm, thick, node distance = 11mm]
\tikzstyle{deterministic}=[diamond, inner sep = 0cm, minimum size = 10mm, thick, draw =black!80, node distance = 10mm]
\tikzstyle{connect}=[-latex, thick]
\tikzstyle{box}=[rectangle, draw=black!100]

\node[blank] (X1) [] {};
\node[blank] (X2) [right=of X1] {};
\node[main,fill=black!10] (X3) [right = 0.05cm of X2] {$\xx$};
\node[main] (A1) [below=of X1] {$a_1$};
\node[main] (A2) [right=of A1] {$a_2$};
\node[main] (A3) [right=of A2] {$a_3$};
\node[main] (A4) [right=of A3] {$a_4$};

\node[deterministic] (H1) [below=of A1] {$\dech_1$};
\node[deterministic] (H2) [right=of H1,below=of A2] {$\dech_2$};
\node[deterministic] (H3) [right=of H2,below=of A3] {$\dech_3$};
\node[deterministic] (H4) [right=of H3,below=of A4] {$\dech_4$};

\node[main,fill=black!10] (Y1) [below=of H1] {$\yy_1$};
\node[main,fill=black!10] (Y2) [right=of Y1,below=of H2] {$\yy_2$};
\node[main,fill=black!10] (Y3) [right=of Y2,below=of H3] {$\yy_3$};
\node[main,fill=black!10] (Y4) [right=of Y3,below=of H4] {$\yy_4$};

\path (A1) edge [connect] (A2);
\path (A2) edge [connect] (A3);
\path (A3) edge [connect] (A4);

\path (H1) edge [connect] (Y1);
\path (Y1) edge [connect] (H2);
\path (H1) edge [connect] (H2);
\path (H1) edge [connect] (A2);
\path (A1) edge [connect,bend left] (Y1);

\path (H2) edge [connect] (Y2);
\path (Y2) edge [connect] (H3);
\path (H2) edge [connect] (H3);
\path (A2) edge [connect,bend left] (Y2);

\path (H3) edge [connect] (Y3);
\path (A3) edge [connect,bend left] (Y3);
\path (H2) edge [connect] (A3);
\path (H3) edge [connect] (A4);
\path (H3) edge [connect] (H4);
\path (Y3) edge [connect] (H4);
\path (A4) edge [connect,bend left] (Y4);
\path (H4) edge [connect] (Y4);

\path (X3) edge [connect] (A1);
\path (X3) edge [connect] (A2);
\path (X3) edge [connect] (A3);
\path (X3) edge [connect] (A4);

\end{tikzpicture}
\end{adjustbox}
\caption{Our monotonic hard-attention model viewed as a graphical model.
The circular nodes are random variables and
the diamond nodes are deterministic variables. 
We have omitted
arcs from $\xx$ to $y_1$, $y_2$, $y_3$ and $y_4$ for clarity (to avoid crossing arcs).
}
\label{fig:graphical-model}
\vspace{-.5cm}
\end{figure}

\paragraph{A Note on $\eos$.}
In the discussion above, we have 
suppressed the generation of $\eos$ in the autoregressive models we derive for brevity.
For example, $p(y_i \mid a_i, \yy_{< i}, \xx)$ must be a conditional distribution over $\Sigmay \cup \{\eos\}$ in order for $p(\yy \mid \xx)$ to be a well-defined probability distribution.\looseness=-1 

\section{A Neural Parameterization with Enforced Monotonicity}
The goal of this section is to take the
$1^\text{st}$-order model of \cref{sec:hard-attention} and show how we can straight-forwardly enforce the monotonicity of the alignments. We will achieve this by adding structural zeros to the distribution, which will still allow us to perform efficient inference with dynamic programming.
We follow the neural parameterization of \newcite{wu2018hard}. The source string $\xx$ is represented by a sequence of character embedding vectors, which are fed into
an encoder bidirectional LSTM
\citep{hochreiter1997long} to produce
hidden state representations 
$\ench_j $. The emission distribution $p(y_i \mid a_i, \yy_{< i}, \xx)$ depends
on these encodings $\ench_j$ and the decoder hidden states $\dech_i$, produced by
\begin{align}
\dech_i &= \lstm([\dece(y_{i-1}) ; \tagh], \dech_{i-1})
\end{align}
where $\dece$ encodes target characters into character embeddings. The tag embedding $\tagh$ is produced by
\begin{align}
\tagh = \relu(\mathbf{Y}\,[\tage(\ttt_1);\dots;\tage(\ttt_{|\Sigmat|})])
\end{align}
where $\tage$ maps the tag $\ttt_k$ into tag embedding $\tagh_k \in \mathbb{R}^{d_t}$
or zero vector $\mathbf{0} \in \mathbb{R}^{d_t}$, depends on whether the tag
$\ttt_k$ is presented. Note that $\mathbf{Y} \in \mathbb{R}^{d_t \times
|\Sigmat|\,d_t}$ is a learned parameter matrix. 
Also, $\ench_{a_i}\in \mathbb{R}^{2d_h}$, $\dech_{a_i} \in \mathbb{R}^{d_h}$ and $\tagh\in \mathbb{R}^{d_t}$ are hidden states. 

\paragraph{The Emission Distribution.}
All of our hard-attention models employ the same 
emission distribution parameterization, which we define below
\begin{align}
p(y_i \mid a_i, \yy_{< i}, \xx) &= \softmax \left(\mathbf{W} \,\NN(\dech_i, \ench_{a_i} ) \right)_{y_i} \nonumber \\
\NN(\dech_i, \ench_{a_i} ) &= \tanh \left(\mathbf{V} \,[\dech_i; \ench_{a_i} ] \right)
\end{align}
where $\mathbf{V} \in \mathbb{R}^{3d_h \times 3d_h}$ and $\mathbf{W} \in \mathbb{R}^{|\Sigmay| \times 3d_h}$ are learned parameters.

\paragraph{0$^\text{th}$-order Hard Attention.} In
the case of the 0$^\text{th}$-order model, the distribution is computed
by a bilinear attention function with \Cref{eq:0th-order}
\begin{equation}
\begin{split}
p(a_i \mid \yy_{<i}, \xx) = \dfrac{\exp({\dech_{i}}^{\top}\, \mathbf{T}\, \ench_{a_i})}{\sum_{j'=1}^{|\xx|} \exp({\dech_{i}}^{\top}\, \mathbf{T}\, \ench_{j'})}
\end{split}
\end{equation}
where $\mathbf{T} \in \mathbb{R}^{d_h \times 2d_h}$ is a learned parameter and $A_i$ is a random variable range over the values of the $i^{\text{th}}$ alignment.\looseness=-1

\paragraph{0$^\text{th}$-order Hard Monotonic Attention.}
We 
may enforce
string monotonicity by zeroing out any non-monotonic alignment \emph{without} adding any additional parameters, which can be done by adding structural zeros to the distribution
as follows\looseness=-1
\begin{equation}
\begin{split}
p(&a_i \mid a_{i-1}, \yy_{<i}, \xx) \\
&=\frac{ \mathbbm{1}\{a_i\geq a_{i-1}\} \exp({\dech_{i}}^{\top}\, \mathbf{T}\, \ench_{a_i})}{\sum_{j'=1}^{|\xx|} \mathbbm{1}\{j'\geq a_{i-1}\} \exp({\dech_{i}}^{\top}\, \mathbf{T}\, \ench_{j'})}
\end{split}
\end{equation}
These structural zeros prevent the alignments from jumping backwards during transduction and, thus,
enforce monotonicity. The parameterization is identical to the $0^\text{th}$-order model up to
the enforcement of the hard constraint with \cref{eq:1st-order}.

\paragraph{1$^\text{st}$-order Hard Monotonic Attention.} 
We may also generalize the $0^\text{th}$-order case by \emph{adding more parameters}. This will equip the model with a more expressive
transition function. In this case, we
take the 1$^\text{st}$-order hard attention to  be an offset-based
transition distribution similar to \newcite{wang2018neural}:\looseness=-1
\begin{align}
&p(a_i\mid a_{i-1}, \yy_{<i}, \xx) \\
&=\begin{cases}
\softmax(\mathbf{U} [\dech_i ; \mathbf{T}\, \ench_{a_{i-1}}] )_{a_i} & 0 \leq\Delta \leq w  \\
0 & \textbf{otherwise }
\end{cases} \nonumber
\end{align}
where $\Delta = a_i - a_{i-1}$ is relative distance to previous attention position, $\mathbf{U} \in \mathbb{R}^{(w+1) \times 2d_h}$ is a learned parameter, and $w \in \mathbb{N}$ is an integer hyperparameter.
Note that, as before, we also enforce monotonicity as
a hard constraint in this parameterization.\looseness=-1

\section{Related Work}
There have been previous attempts
to look at monotonicity in neural 
transduction.
\newcite{graves2012sequence} first
introduced the monotonic neural transducer 
for speech recognition.
Building on this, \newcite{yu2016online} proposes using a separated \texttt{shift}/\texttt{emit} transition distribution to allow a more expressive model. Like us, they also consider morphological inflection and outperform a (weaker) soft attention baseline.
\newcite{rastogi-cotterell-eisner:2016:N16-1} offer
a neural parameterization of a
finite-state transducer, which implicitly encodes
monotonic alignments. 
Instead of learning the alignments directly,
\newcite{aharoni-goldberg:2017:Long} take the monotonic alignments from an
external model \citep{sudoh2013noise} and train the neural model with these alignments. In follow-up work, \newcite{makarov2017align} show this two-stage 
approach to be effective, winning the CoNLL-SIGMORPHON 2017 shared task on morphological inflection
\citep{cotterell-conll-sigmorphon2017}. \newcite{raffel2017online} propose a
stochastic monotonic transition process to allow sample-based \emph{online} decoding.\looseness=-1

\section{Experiments}

\subsection{Experiments Design}

\paragraph{Tasks.} We consider three character-level transduction tasks:
grapheme-to-phoneme conversion
\citep{CMUDict,Sejnowski1987ParallelNT}, named-entity transliteration
\citep{Zhang2015WhitepaperON} and morphological inflection in high-resource setting \citep{cotterell-conll-sigmorphon2017}.

\paragraph{Empirical Comparison.} We compare (i) soft attention without input-feeding (\textsc{soft})
\citep{Luong2015EffectiveAT}, (ii) 0$^\text{th}$-order hard attention (\textsc{0-hard})
\citep{wu2018hard}, (iii) 0$^\text{th}$-order monotonic hard attention (\textsc{0-mono})
and (iv) 1$^\text{st}$-order monotonic hard attention (\textsc{1-mono}). 
The \textsc{soft}, \textsc{0-hard} and \textsc{0-mono} models have an \emph{identical} number of
parameters, but the \textsc{1-mono} has more.  All of them have approximately 8.6M parameters.
Experimental details and hyperparameters may be found in \cref{appendix:detail}.

\subsection{Experimental Findings}

\paragraph{Finding \#1: Morphological Inflection.} 
The first empirical finding in our study is that
we achieve single-model, state-of-the-art performance on the CoNLL-SIGMORPHON 2017 shared
task dataset. The results are shown in \cref{table:all}. We find that the \textsc{1-mono} ties with the \textsc{0-mono} system, indicating the additional parameters do not add much. Both of these monotonic systems surpass the non-monotonic system \textsc{0-hard} and \textsc{soft}. 
We also compare to other top systems at the task in \cref{table:decode}. The previous state-of-the-art model, \newcite{bergmanis-EtAl:2017:K17-20}, is a non-monotonic system that outperformed the monotonic system of \newcite{makarov2017align}. 
However,
\newcite{makarov2017align} is a pipeline system that took alignments from an existing aligner; such a system has no manner, by which it can recover from poor initial alignment. We show that \emph{jointly} learning monotonic alignments leads to improved results.%

\paragraph{Finding \#2: Effect of Strict Monotonicity.} 
The second finding is that 
by comparing \textsc{soft}, \textsc{0-hard},
\textsc{0-mono} in \cref{table:all}, we observe
\textsc{0-mono} outperforms \textsc{0-hard} and \textsc{0-hard} in turns outperforms \textsc{soft} in all three tasks.
This shows that monotonicity should be enforced
strictly since strict monotonicity does not hurt the model. 
We contrast this to the findings of \newcite{wu2018hard}, who found the non-monotonic models outperform the monotonic ones; this suggests strict monotonicity is more helpful when the model is allowed to learn
the alignment distribution jointly.\looseness=-1

\begin{table}
\centering
 \begin{adjustbox}{width=.7\columnwidth}
\begin{tabular}{ll} \toprule
Morphological Inflection & \textbf{ACC} \\ \midrule
\newcite{silfverberg-EtAl:2017:K17-20} & 93.0 \\
\textsc{soft} & 93.4 \\
\newcite{makarov2017align}  & 93.9 \\
\textsc{0-hard} & 94.5 \\
\newcite{bergmanis-EtAl:2017:K17-20}  & 94.6\\
\newcite{makarov2018imitation} & 94.6 \\
\textsc{0-mono} & \textbf{94.8} \\
\textsc{1-mono} & \textbf{94.8} \\
\bottomrule
\end{tabular}
\end{adjustbox}
\caption{Average dev performance on morphological inflection of our models against single models
from the 2017 shared task. All systems are single model, i.e., \emph{without} ensembling. Why dev? No participants submitted single-model systems for evaluation on test and the best systems were not open-sourced, constraining our comparison. Note we report numbers from their paper.\protect\footnotemark}
\label{table:decode}
\end{table}

\footnotetext{Some numbers were obtained by contacting authors.}

\paragraph{Finding \#3: Do Additional Parameters Help?}
The third finding is that \textsc{1-mono} has a more expressive transition distribution and, thus, outperforms \textsc{0-mono} and \textsc{0-hard} in G2P. However, it performs as well as or worse on the other tasks. 
This tells us that the additional parameters are not always necessary for improved performance. 
Rather, it is the hard constraint that matters---not the more expressive distribution. 
However, we remark that enforcing the monotonic constraint does come at an additional computational cost.\looseness=-1

\section{Conclusion}
We expand the hard-attention neural sequence-to-sequence model of \newcite{wu2018hard} to enforce monotonicity. We show, empirically, that enforcing monotonicity in the alignments found by hard attention models helps significantly, and we achieve state-of-the-art performance
on the morphological inflection using data from the CoNLL-SIGMORPHON 2017 shared task.  We isolate the effect of monotonicity in a controlled experiment and show monotonicity is a useful hard constraint for three tasks, and speculate previous underperformance is due to a lack of joint training. 

\begin{table}
\centering
\begin{adjustbox}{width=\columnwidth}
\begin{tabular}{l ll ll ll} \toprule
& \multicolumn{2}{l}{Trans} & \multicolumn{2}{l}{G2P} & \multicolumn{2}{l}{MorInf} \\
\cmidrule(lr){2-3} \cmidrule(lr){4-5} \cmidrule(lr){6-7}
& \textbf{ACC} & \textbf{MFS} & \textbf{WER} & \textbf{PER} & \textbf{ACC} & \textbf{MLD} \\ \midrule \vspace{.1cm}
\textsc{soft}     & 40.4 & 0.893 & 29.3 & 0.071 & 92.9 & 0.157 \\
\textsc{0-hard}    & 41.1$^{\star}$ & 0.894 & 29.2$^{\star}$ & {0.070} & 93.8$^{\star}$ & 0.126 \\
\textsc{0-mono}  & \textbf{41.2}$^{\star}$ & \textbf{0.895} & {29.0}$^{\star \times}$ & 0.072 & \textbf{94.4}$^{\star \times}$ & \textbf{0.113} \\
\midrule
\textsc{1-mono} & 40.8 & 0.893 & \textbf{28.2}$^{\star \times \dagger}$ & \textbf{0.069} & \textbf{94.4}$^{\star \times}$ & 0.116 \\
\bottomrule
\end{tabular}
\end{adjustbox}
\caption{Average test performance of namded-entity transliteration (Trans), grapheme-to-phoneme conversion (G2P) and morphological inflection (MorInf).
First group has exactly same number of parameter while the second group has slightly more parameter.
$\star$, $\times$ and $\dagger$ indicate statistical significant improvement against \textsc{soft}, \textsc{0-hard} and \textsc{0-mono} on language-level paired permutation test ($p < 0.05$).
}
\label{table:all}
\end{table}

\section*{Acknowledgments}
RC acknowledges a Facebook Fellowship.\looseness=-1

\bibliographystyle{acl_natbib}
\bibliography{main}

\cleardoublepage
\appendix

\onecolumn
\section{Experimental Details}\label{appendix:detail}

\subsection{Tasks}
We use \citeposs{wu2018hard} data splits of
grapheme-to-phoneme conversion \cite[CMUDict;][]{CMUDict} and NetTalk
\cite{Sejnowski1987ParallelNT} and NEWS 2015 shared task on named-entity
transliteration. 
In named-entity transliteration, we only run experiments on 11
language pairs.\footnote{Ar–En, En–Ba, En–Hi, En–Ja, En–Ka, En–Ko, En–Pe, En–Ta,
En–Th, Jn–Jk and Th–En.}
Grapheme-to-phoneme conversion is evaluated by word error rate (WER) and phoneme
error rate \cite[PER;][]{DBLP:conf/interspeech/YaoZ15}, where PER is the edit
distance divided by the number of the phonemes. 
Named-entity transliteration is
evaluated by word accuracy (ACC) and mean F-score 
\citep[MFS;][]{Zhang2015WhitepaperON}. F-score is computed by\looseness=-1
\begin{subequations}
\begin{align}
\text{LCS}(c_i, r_i) &= \frac{1}{2} (|c_i|+|r_i|-\text{ED}(c_i,r_i)) \\
R_i &= \frac{\text{LCS}(c_i, r_i)}{|r_i|} \\
P_i &= \frac{\text{LCS}(c_i, r_i)}{|c_i|} \\
\text{FS}_i &= 2 \frac{R_i\times P_i}{R_i + P_i}
\end{align}
\end{subequations}
where $r_i$ and $c_i$ is the $i^\text{th}$ reference and prediction and $\text{ED}(c_i, r_i)$ is the
edit distance between $c_i$ and $r_i$. Morphological inflection is evaluated by word
accuracy (ACC) and average edit distance (MLD)
\cite{cotterell-conll-sigmorphon2017}.\looseness=-1

\subsection{Parameterization}\label{subsec:param}

For completeness, we also include the parameterization of transducer with soft attention:
\begin{subequations}
\begin{align}
p(y_i \mid \yy_{< i}, \xx) &= \softmax \left(\mathbf{W} \,\NN(\dech_i, \cc_i ) \right)_{y_i} \\
\cc_i &= \sum_{j=1}^{|\xx|} \alpha_{ij} \, \ench_j \\
e_{ij} &= {\dech_i}^\top\,\mathbf{T}\,\ench_j \\
\alpha_{ij} &= \frac{\exp(e_{ij})}{\sum^{|\xx|}_{j'=1} \exp(e_{ij'})} 
\end{align}
\end{subequations}
The dimensions of the character and tag embedding are 200 and 40, respectively. 
The
encoder and decoder LSTM both have 400 hidden dimensions ($d_h$). We also have a 2 layer
encoder LSTM. We have 0.4 dropout in embedding and encoder LSTM. The $w$ in 1$^\text{st}$-order hard monotonic attention model is 4.

\subsection{Optimization}
The model is trained with Adam \citep{kingma2014adam} and the learning rate is
0.001. 
We halve the learning rate whenever the development log-likelihood
increase and we stop early when the learning rate reaches 0.00001. 
We apply
gradient clipping with maximum gradient norm 5. The models are selected by
development evaluation metric and decoded greedily since no improvements are
observed when using beam search \citep{wu2018hard}.

\end{document}